# Infrared Small Target Detection Using Double-Weighted Multi-Granularity Patch Tensor Model With Tensor-Train Decomposition

Guiyu Zhang, *Student Member, IEEE*, Qunbo Lv, *Member, IEEE*, Zui Tao, Baoyu Zhu, *Member, IEEE*, Zheng Tan, *Member, IEEE*, Yuan Ma

*Abstract*—Infrared small target detection plays an important role in the remote sensing fields. Therefore, many detection algorithms have been proposed, in which the infrared patch-tensor (IPT) model has become a mainstream tool due to its excellent performance. However, most IPT-based methods face great challenges, such as inaccurate measure of the tensor low-rankness and poor robustness to complex scenes, which will lead to poor detection performance. In order to solve these problems, this paper proposes a novel double-weighted multi-granularity infrared patch tensor (DWMGIPT) model. First, to capture different granularity information of tensor from multiple modes, a multi-granularity infrared patch tensor (MGIPT) model is constructed by collecting nonoverlapping patches and tensor augmentation based on the tensor train (TT) decomposition. Second, to explore the latent structure of tensor more efficiently, we utilize the auto-weighted mechanism to balance the importance of information at different granularity. Then, the steering kernel (SK) is employed to extract local structure prior, which suppresses background interference such as strong edges and noise. Finally, an efficient optimization algorithm based on the alternating direction method of multipliers (ADMM) is presented to solve the model. Extensive experiments in various challenging scenes show that the proposed algorithm is robust to noise and different scenes. Compared with the other eight state-of-the-art methods, different evaluation metrics demonstrate that our method achieves better detection performance in various complex scenes.

## I. INTRODUCTION

Infrared search and track (IRST) system achieves the discrimination and recognition of the interested target based on the characteristic differences between target thermal radiation and background thermal radiation. On the one hand, compared with visible light imaging, IRST system has better smoke penetration capability [1]. On the other hand, compared with active radar imaging, IRST system has better concealment [2]. It is indispensable in aerospace technology [3], remote surveillance [4], target tracking [5]. However, infrared small target detection still faces many challenges, mainly reflected in the following aspects. First, due to limitations of the long imaging distance and sensor pixel size, the target size is small in infrared images, resulting in a lack of structural and textural information [6]. Second, interfered by undulating background clutter such as clouds, sea, buildings and system noise, the contrast between the target and the background is usually low, which will lead to false detection [7]. In addition, the movement of the target appears randomly without specific rules, and there are some false alarm sources similar to the characteristics of the target in image, which increases the difficulty of target search, detection and tracking [8]. Therefore, infrared small target detection in complex scenes is still a great challenge.

In past decades, aiming at high accuracy and strong robustness, a variety of infrared small target detection algorithms have been proposed, which can be classified into two categories: single-frame image-based detect-before-track (DBT) and sequence-based track-before-detect (TBD) [9]. TBD aligns multi-frame information to achieve target energy accumulation, such as 3D matched filtering [10], dynamic programming algorithm [11], spatiotemporal saliency approach [12], and particle filtering [13]. Nevertheless, it is difficult to ensure the consistency of the background in practical application owing to factors such as rapid movement of the observed target and sensor jitter. As a result, TBD has poor performance in cloud, sea and other scenes where multi-frame alignment is not easily achievable. At the same time, TBD has high requirements for storage and computing resources. Compared with TBD, DBT requires less initial assumptions and prior knowledge, and has the characteristics of high detection accuracy, fast running speed and wide application scenarios, which has become a research hotspot in

This work was supported in part by the Key Program Project of Science and Technology Innovation of Chinese Academy of Sciences (no. KGFZD-135-20-03-02) and Innovation Foundation of Key Laboratory of Computational Optical Imaging Technology, CAS (no. CXJJ-23S016). *(Corresponding authors: Qunbo Lv; Zheng Tan.)*

Guiyu Zhang, Qunbo Lv, Baoyu Zhu, and Zheng Tan are with the College of Optoelectronics, University of Chinese Academy of Sciences, Beijing 100049, China, and also with the Aerospace Information Research Institute, Chinese Academy of Sciences, Beijing 100094, China (e-mail: zhangguiyu21@mails.ucas.ac.cn; lvqunbo@aoe.ac.cn; zhubaoyu20@mails.ucas.ac.cn; tanzheng@aircas.ac.cn).

Zui Tao is with the Aerospace Information Research Institute, Chinese Academy of Sciences, Beijing 100101, China (e-mail: taozui@radi.ac.cn).

Yuan Ma is with the Department of Mechanical Engineering, Tsinghua University, Beijing 100084, China (yuanma@tsinghua.edu.cn).

recent years [14]. DBT can be divided into three categories: 1) filter-based methods; 2) human visual system (HVS)-based methods; 3) low-rank and sparse decomposition-based methods.

*A. Related Work*

Filter-based methods enhance the target features based on the prior information of background consistency. Typical methods include Top-hat filter [15], Max-Mean and Max-Median filters [16], and high-pass filter [17]. Hadhoud and Thomas [18] extended the LMS algorithm [19] and proposed a two-dimensional adaptive least mean square (TDLMS) filter. Cao and Sun [20] utilized the maximum inter-class variance method to improve morphological filtering. Sun et al. [21] proposed a novel spatiotemporal filtering combined with directional filter bank (DFB). However, such methods usually require the background to be continuous, thus their performance degrades significantly when faced with complex scenes full of edges and noise.

According to neuro-physiological findings, contrast is the most crucial factor encoded in our visual system, which is also of great importance in the detection process [22]. Based on this, Chen et al. [23] proposed a local contrast method (LCM) to describe the difference between the point and its neighborhood. Inspired by LCM, many methods based on local contrast improvement are proposed. Han et al. [24] improved the detection speed by increasing the sliding window step size and proposed an improved local contrast measure (ILCM). Starting from the perspective of image patch difference, Wei et al. [25] proposed a multiscale patch-based contrast measure (MPCM). Shi et al. [26] proposed a high-boost-based multiscale local contrast measure (HBMLCM). Han et al. [27] proposed a multiscale tri-layer local contrast measure (TLLCM) for computing comprehensive contrast. However, when the image contains background edges and pixel-sized noises with high brightness (PNHB), such algorithms usually have difficulty in distinguishing between target, background and noise, leading to false detection.

Recently, low-rank and sparse decomposition-based methods have achieved great success, which can effectively separate the low rank background component and the sparse target component of infrared image. Generally, the infrared image can be modeled as linear superposition of target image, background image, and noise image:

$$f_D = f_B + f_T + f_N \qquad (1)$$

where $f_D, f_B, f_T$ and $f_N$ represent the original infrared image, background image, target image and noise image, respectively. Gao et al. [28] first extended the infrared image model to the infrared patch-image (IPI) model through local patch construction, and constrained the low-rank background component and sparse target component with nuclear norm minimization (NNM) and $l_1$-norm respectively, thus transforming the infrared small target detection into an optimization problem. However, as the NNM uses the same threshold to shrink singular values, over-shrinkage problem may occur in complex backgrounds full of interference [29]. Furthermore, besides the target, edges and corners in the background are also considered as sparse component under $l_1$-norm [30]. To handle the above problems, Dai et al. [31] constructed a non-negative infrared patch-image (NIPPS) model by adding a non-negative constraint to sparse target component. Wang et al. [32] proposed a stable multi-subspace learning (SMSL) model by assuming that the background data comes from a mixture of low-rank subspaces and constrains each subspace with row-1 norm. Wang et al. [33] appended an additional total variation regularization term to low-rank background component, and proposed total variation regularization and principal component pursuit (TV-PCP). Zhang et al. [34] proposed nonconvex rank approximation minimization (NRAM) by utilizing the $l_{2,1}$-norm to constrain remaining edges. Zhang et al. [35] combined $l_p$-norm to constrain sparse target component and proposed non-convex optimization with $l_p$-norm constraint (NOLC).

In order to exploit image structure and reduce computing cost [36], Dai and Wu [37] extended IPI to the third-order tensor domain and proposed reweighted infrared patch-tensor (RIPT). Zhang and Peng [38] utilized the partial sum of tensor nuclear norm (PSTNN) to approximate the tensor rank, which effectively reduces computational time. Kong et al. [39] proposed a nonconvex tensor fibered rank approximation (NTFRA) method, which uses the tensor fibered nuclear norm based on the Log operator (LogTFNN) to nonconvex approximate the tensor fibered rank and removes noise with the help of hypertotal variation (HTV) as a joint regularization term. Zhang et al. [40] constructed a non-local block tensor and a adaptive compromising factor based on the image local entropy. Then, a self-adaptive and non-local patch-tensor (ANLPT) model was proposed for infrared small target detection.

*B. Motivation*

Compared with filter-based approaches and HVS-based approaches, tensor-based methods can better enhance small target features and suppress background clutter interference. However, due to the complex multilinear structure of the tensor, the exact approximation of the tensor rank is always a major difficulty. Therefore, many scientists focus on selecting more accurate tensor-rank constraint. The Tucker rank [41] was firstly applied to the RIPT model, with the employment of the sum of nuclear norm (SNN) as its convex approximation. However, Tucker rank cannot appropriately capture the global correlation of high-dimensional data [42] and SNN is just a suboptimal approximation of Tucker rank minimization [43]. Then, the tensor tubal rank and its convex surrogate tensor nuclear norm (TNN) [44] based on the tensor singular value decomposition(t-SVD) [45] was proposed, which is applied to approximate the tensor rank in PSTNN. According to the definition of TNN, the correlation of high-dimensional data is only characterized by a single mode based on the t-SVD, which implies that the TNN lacks flexibility and a measure of the low-rankness from multiple modes. In order to alleviate the shortage of TNN, in NTFRA the tensor rank is expanded to the tensor fibered rank [46] whose nonconvex surrogate is LogTFNN based on the multimodal t-SVD. In summary, the prior assumption of IPT-based models based on the above tensor-rank constraints is that the patch space is a third-order tensor, and the unfolding matrix dimension of a patch-tensor mainly depends on the patch size when obtaining information from nonlocal self-correlation in tensor space. Since the main difficulty of infrared small target detection is a lack of sufficient information, the above matricization strategies which only relies on patch size is inflexible and insufficient to deal with highly complex scenes.

In addition, sparse interference such as background edges with strong contrast and PNHB is also a main reason for false alarms in the tensor-based methods. In recent years, local structure tensor [37]-[39] has been widely used in IPT-based models. However, this method cannot effectively suppress edge information in complex scenes, leading to false alarms. In order to solve these problems and further improve the performance of infrared small target detection in complex scenes, we innovatively proposed an infrared small target detection method based on double-weighted multi-granularity infrared patch tensor (DWMGIPT) model. First, in order to capture more information about image structure and estimate tensor rank more accurately, we employ TT decomposition to obtain multi-granularity information and use an auto-weighted mechanism to evaluate the importance of different granularity information, further excavate the internal relationship between the data. Second, as the steering kernel (SK) is strongly robust to noise interference and can reflect the internal structure of the image [47]-[50], the target and the background can be more effectively separated by combining the SK-based local structure prior and the TT decomposition. Experimental results show that the proposed algorithm outperforms other state-of-the-art methods in terms of accuracy and robustness in various complex scenes. The main contributions of this paper are as follows.

1) We propose a novel multi-granularity infrared patch tensor (MGIPT) model by using Tensor-Train (TT) Nuclear Norm as the convex surrogate of TT rank, which can extract different granularity information and obtain accurate background estimation.
2) The auto-weighted mechanism is used to assess the significance of multi-granularity information under different matricization modes. Meanwhile, we propose a new method for calculating local structure prior based on SK, which can enhance the target and suppress the background clutter.
3) We apply auto-weighted mechanism and SK-based local structure prior to MGIPT model for infrared small target detection and introduce an efficient optimization scheme based on the alternating direction method of multipliers (ADMM).

The rest of this article is organized as follows. Section II introduces basic notations and preliminary knowledge. In Section III, we provide the theoretical analysis and construct an optimization algorithm of the proposed model. Section IV conducts extensive experiments to verify the detection performance of the proposed algorithm in various complex scenes. Section V summarizes this article and discusses the future work.

II. NOTATIONS AND PRELIMINARIES

*A. Notations*

In this article, the notations we adopt are defined as follows. Scalars, vectors and matrices are denoted by lowercase letters (e.g., $x$, $x \in \mathbb{R}$), boldface lowercase letters (e.g., $\mathbf{x}$, $\mathbf{x} \in \mathbb{R}^I$) and capital letters (e.g., $X$, $X \in \mathbb{R}^{I \times J}$), respectively. Considering tensors are multi-index arrays, we use Euler scrip (e.g., $\mathcal{X}$, $\mathcal{X} \in \mathbb{R}^{I_1 \times I_2 \times \cdots \times I_N}$) to represent $N$th-order tensor ($N \geq 3$). An element in the tensor $\mathcal{X} \in \mathbb{R}^{I_1 \times I_2 \times \cdots \times I_N}$ is represented as $\mathcal{X}(i_1, i_2, \cdots, i_N)$ or $x_{i_1,i_2,\cdots,i_N}$, where $(i_1, i_2, \cdots, i_N)$ is the index. The mode-i unfolding operation for TT decomposition is denoted as $X_{[i]} = \text{reshape}_{[i]}(\mathcal{X}) \in \mathbb{R}^{(\prod_{j=1}^{i} I_j) \times (\prod_{j=i+1}^{N} I_j)}$ and the corresponding matrix-tensor folding is denoted as $\mathcal{X} = \text{unreshape}_{[i]}(X_{[i]})$. The $l_0$-norm of a tensor is defined as number of non-zero elements of the tensor, the $l_1$-norm is defined as $\|\mathcal{X}\|_1 = \sum_{i_1,i_2,\cdots,i_N} |x_{i_1,i_2,\cdots,i_N}|$, and the Frobenius norm is defined as $\|\mathcal{X}\|_F = \sqrt{\sum_{i_1,i_2,\cdots,i_N} x_{i_1,i_2,\cdots,i_N}^2}$.

*B. Tensor-Train Decomposition*

The main problem of the tensor-based model is the definition of tensor rank, with several popular definitions are Tucker rank, tubal rank and tensor-train rank (TT rank). Compared with others, the TT rank can capture the global correlation of a tensor as it contains a correlation between a few modes (rather than a single mode) and the rest of the tensor [42]. For example, suppose $N$th-order tensor $\mathcal{X}$ with all the same dimension ($d_1 = d_2 = \cdots = d_N = d$), then matrix $X_{[i]}$ has a dimension of $\prod_{j=1}^{i} d_j \times \prod_{j=i+1}^{N} d_j$. Compared with all unfolding matrices of Tucker rank which have the same dimension of $d \times d^{N-1}$, TT rank has a well-balanced matricization scheme.

Definition 1 (Tensor-Train (TT) Rank) [51]: For a $N$th-order tensor $\mathcal{X} \in \mathbb{R}^{I_1 \times I_2 \times \cdots \times I_N}$, the TT rank is defined as:

$$\text{rank}_{tt}(\mathcal{X}) = (\text{rank}(X_{[1]}), \text{rank}(X_{[2]}), \cdots, \text{rank}(X_{[N-1]})) \quad (2)$$

where $X_{[i]} \in \mathbb{R}^{(\prod_{j=1}^{i} I_j) \times (\prod_{j=i+1}^{N} I_j)}$ is the mode-i unfolding of $\mathcal{X}$.

Definition 2 (Tensor-Train (TT) Nuclear Norm) [42]: The tensor-train nuclear norm (TTNN) of tensor $\mathcal{X} \in \mathbb{R}^{I_1 \times I_2 \times \cdots \times I_N}$ as the convex surrogate of the TT rank, is defined by the weighted sum of different unfolding matrices:

$$\|\mathcal{X}\|_* = \sum_{i=1}^{N-1} \alpha_i \|X_{[i]}\|_* \quad (3)$$

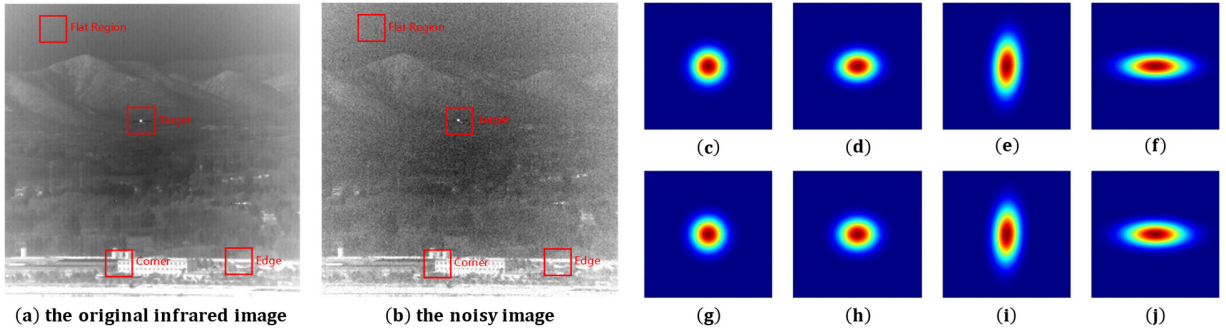

(a) the original infrared image  (b) the noisy image  (c)  (d)  (e)  (f)  (g)  (h)  (i)  (j)

Fig. 1. Robust representation of SK with covariance matrices $\{C_i\}$ for different structures of infrared image. (a) Complex infrared image. (b) The image with Gaussian noise of $\sigma = 10$. (c) - (f) The SK representations of the flat region, target region, corner region and edge region in original infrared image, respectively. (g) - (j) The SK representations of the flat region, target region, corner region and edge region in noisy infrared image, respectively.

where $\|\cdot\|_*$ is the matrix nuclear norm, $\alpha_i$ is the weight of mode-i unfolding and $\sum_{i=1}^{N-1}\alpha_i = 1, \alpha_i > 0$.

### C. Auto-weighted Mechanism

The importance of TT rank of different modes varies and the nuclear norm can be a surrogate for the rank approximately. Therefore, for the following optimization problem:

$$\min_{\mathcal{X}} \ \sum_{i=1}^{N-1} \alpha_i \|X_{[i]}\|_* \quad \text{s.t.} \ \boldsymbol{\alpha}^T \mathbf{1} = 1, \boldsymbol{\alpha} \geq 0 \quad (4)$$

How to choose $\alpha_i$ to measure the importance of TT rank of different modes is necessary. Considering the particularity of matrix rank, the larger rank contains more information, which should deserve larger weight. In order to automatically assign weights according to the matrix rank, Chen et al. [52] introduced an auto-weighted mechanism, where the objective function is expressed as follow:

$$\min_{\boldsymbol{\alpha}} \ -\boldsymbol{\mu}^T\boldsymbol{\alpha} + \psi\|\boldsymbol{\alpha}\|_2^2 \quad \text{s.t.} \ \boldsymbol{\alpha}^T \mathbf{1} = 1, \boldsymbol{\alpha} \geq 0 \quad (5)$$

where vector $\boldsymbol{\mu}^T = \left(\|X_{[1]}\|_*, \|X_{[2]}\|_*, \cdots, \|X_{[N-1]}\|_*\right)$ and $\psi$ is a penalty factor used to smoothen the weight distribution. Obviously, it's a convex Quadratic Programming (QP) and owns a global optimal solution:

$$\alpha_i = \begin{cases} \frac{\mu_i - \eta}{2\psi}, & \mu_i - \eta > 0 \\ 0, & \mu_i - \eta \leq 0 \end{cases} \quad (6)$$

where $\eta = (\sum_{i=1}^{N-1}\mu_i - 2\psi) / j$ and $j$ represents the number of nonvanishing elements in $\boldsymbol{\alpha}$. The larger the nuclear norm of unfolding matrix $X_{[i]}$ is, the larger weight $\alpha_i$ is forced to maintain more information captured in the current mode.

### D. Steering Kernel

The steering kernel (SK) [53] can robustly estimate the structure information by analyzing gradients and radiometric similarities of pixels in a local window, which is modeled as:

$$K(\mathbf{x}_i, \mathbf{x}_j) = \sqrt{\det(C_j)} \exp\left\{-(\mathbf{x}_i - \mathbf{x}_j)^T C_j (\mathbf{x}_i - \mathbf{x}_j)\right\} \quad (7)$$

where $\mathbf{x}_i$ is the pixel position of interest and $\mathbf{x}_j$ denotes a given location inside the local window centered at $\mathbf{x}_i$. The covariance matrix $C_j$ is estimated by the local gradient matrix for the window $w_j$ centered at $\mathbf{x}_j$. The estimation of the steering matrix $C_j$ plays a crucial role for SK to reliably capture image local structure and is specified as follows. As depicted in Fig. 1, different regions have dramatically different SK representations by intuitively comparing the shapes. It can be seen that the SK representation of the flat region (i.e., (c) and (g) in Fig. 1) is circular, while the SK representations of the corner (i.e., (e) and (i) in Fig. 1) and the edge (i.e., (f) and (j) in Fig. 1) are elongated and rotated along the inflection point and strong edge direction, respectively. The SK representation of the target (i.e., (d) and (h) in Fig. 1) is also elongated and rotated along the direction of the target contour but not as obvious as the corner's and edge's. Moreover, it can be observed that the SK exhibits good robustness under noise interference.

Let $\Omega(\mathbf{x}_j) = \{\mathbf{x}_1, \cdots, \mathbf{x}_j, \cdots, \mathbf{x}_M\}$ represent the coordinate set of $M$ neighboring pixels centered at pixel coordinate $\mathbf{x}_j = [x_j^1, x_j^2]^T$ and the local gradient matrix for the window $w_j$ centered at $\mathbf{x}_j$ can be expressed by:

$$G_j = \begin{bmatrix} \widehat{G}_x(\mathbf{x}_1) & \widehat{G}_y(\mathbf{x}_1) \\ \vdots & \vdots \\ \widehat{G}_x(\mathbf{x}_j) & \widehat{G}_y(\mathbf{x}_j) \\ \vdots & \vdots \\ \widehat{G}_x(\mathbf{x}_M) & \widehat{G}_y(\mathbf{x}_M) \end{bmatrix} \quad (8)$$

where $\widehat{G}_x(\cdot)$ and $\widehat{G}_y(\cdot)$ represent the first derivatives along the horizontal and vertical directions, respectively. Based on the singular value decomposition (SVD) of $G_j$, the stable covariance matrix $C_j$ can be estimated by using a regularized parameter approach:

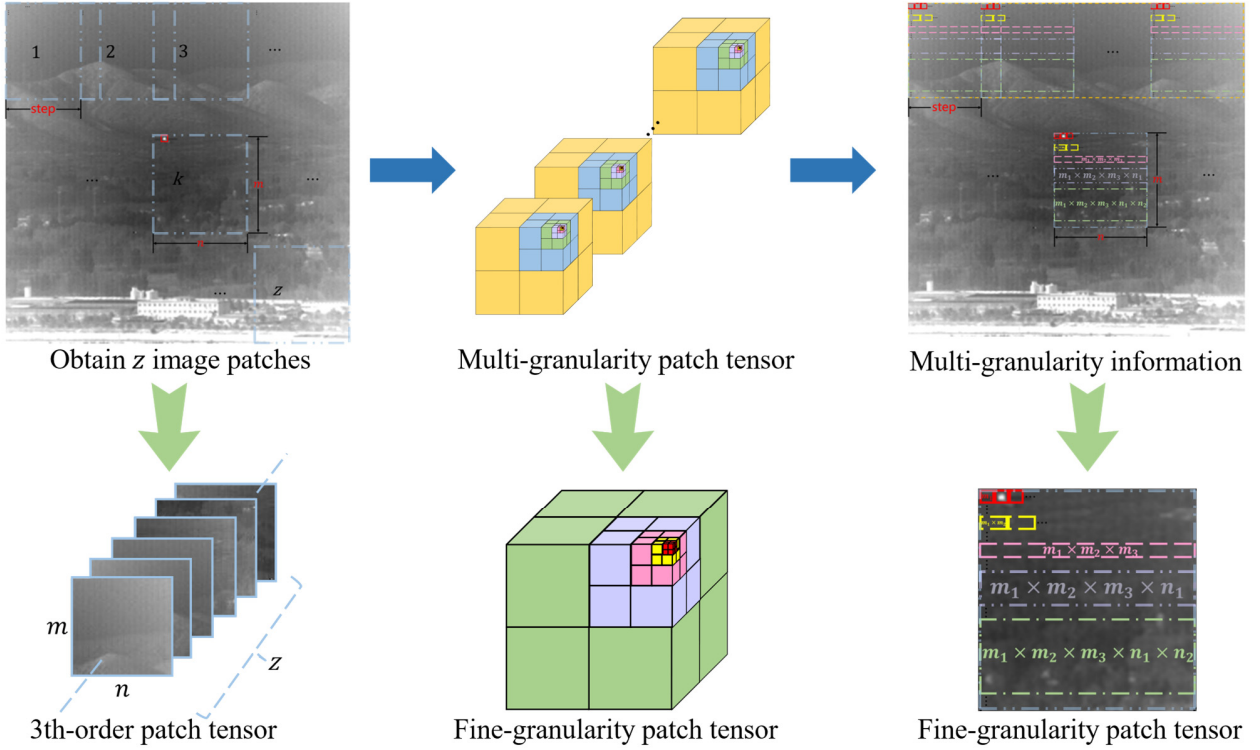

Fig. 2. Multi-Granularity Infrared Patch Tensor model.

$$C_j = \gamma(\tau_1 \mathbf{v}_1 \mathbf{v}_1^T + \tau_2 \mathbf{v}_2 \mathbf{v}_2^T) \qquad (9)$$

with

$$\gamma = \left(\frac{s_1 s_2 + \lambda''}{M}\right)^3, \quad \tau_1 = \frac{s_1 + \lambda'}{s_2 + \lambda'}, \quad \tau_2 = \frac{s_2 + \lambda'}{s_1 + \lambda'} \qquad (10)$$

where $\gamma$ and $\tau_1$ are scaling and elongation parameters, respectively. According to Seo and Milanfar [54], the regularization parameters $\lambda'$ and $\lambda''$ for all experiments are set to 1 and $10^{-7}$, respectively. In addition, the singular values ($s_1 \geq s_2 \geq 0$) and the dominant direction $\mathbf{v}_1$ come from the SVD formula $G_j = U_j S_j V_j^T = U_j \text{diag}[s_1, s_2]_j [\mathbf{v}_1, \mathbf{v}_2]_j^T$.

## II. METHODOLOGY

### A. Multi-Granularity Infrared Patch Tensor Model

The proposed MGIPT model explores different granularity information of tensor, as shown in Fig. 2. Given an infrared image, we first obtain $z$ ($z = cols \times rows$) image patches from top left to bottom right over the image through a sliding window. In order to explore latent structure information of data, we perform prime factorization on patch size $P_{size} \in \mathbb{R}^{m \times n}$ and combine the TT decomposition to decompose the original image patches into different granularity. Finally, the Eq. (1) is transferred to the multi-granularity patch space of tensor:

$$\mathcal{D} = \mathcal{B} + \mathcal{T} + \mathcal{N} \qquad (11)$$

where $\mathcal{D}$, $\mathcal{B}$, $\mathcal{T}$, $\mathcal{N} \in \mathbb{R}^{m_1 \times \cdots \times m_i \times n_1 \times \cdots \times n_j \times cols \times rows}$ are the multi-granularity patch-tensor forms corresponding to model (1). The prime factorization of the height $m$ and width $n$ of the sliding window is $m = m_1 \times \cdots \times m_i$ and $n = n_1 \times \cdots \times n_j$, respectively. Compared with the IPT-based approaches, our data reconstruction model captures different granularity information of the image, providing more views to mine the inner relationship of data.

### B. Local Structure Prior Based on Steering Kernel

The strong edges and corner points in the background are usually sparse and have similar features as the target, making it difficult to distinguish them from the target only by the global sparse features. Fortunately, the above interfering component can be identified using local features, which indicating that the limitations of optimization methods can be mitigated by incorporating local structure prior. Considering that the separate steering matrix estimated at each pixel location can robustly depict the local structure of image, we utilize the two eigenvalues $\lambda_1$ and $\lambda_2 (\lambda_1 \geq \lambda_2)$ of the steering matrix to reflect different local characteristics: at the flat region, $\lambda_1 \approx \lambda_2$; at the corner and the edge region, $\lambda_1 > \lambda_2$. Since the value of $\lambda_1 - \lambda_2$ highlights the image boundary information, we obtain two matrices $L_1$ and $L_2$ by calculating eigenvalues of steering matrix at all pixel locations and define the target prior as follows:

$$W_{\text{tp}} = \exp\left(\frac{(L_1 - L_2) - t_{\min}}{t_{\max} - t_{\min}}\right) \qquad (12)$$

where $t_{max}$ and $t_{min}$ are the maximum and minimum of $L_1 -$

$L_2$, respectively. Fig. 3(a) displays the original image and Fig. 3(b) exhibits the map of target obtained through the Eq. (12). Although the target information is highlighted that fully satisfies our model requirement, it is hard to only utilize the Eq. (12) to distinguish the target from the background due to the similarity between the target edge and the background edge. Therefore, we use maximum operation between two eigenvalues to obtain the background prior:

$$L(x,y) = \max(L_1(x,y), L_2(x,y)) \quad (13)$$

$$W_{bp} = \frac{L - b_{min}}{b_{max} - b_{min}} \quad (14)$$

where $(x, y)$ denotes the pixel position, and $b_{max}$ and $b_{min}$ are the maximum and minimum of $L$, respectively. Fig. 3 (c) shows the background prior, which precisely extracts the background information. In order to distinguish the target from the background more accurately, we consider both the target prior and the background prior to obtain the final prior map:

$$W_p = W_{tp} \odot W_{bp} \quad (15)$$

where $\odot$ represents the Hadamard product. Compared with local structure tensor weights used in RIPT (i.e., Fig. 3(d)) and PSTNN (i.e., Fig. 3(e)), we can observe that while the strong edges are not completely eliminated, our prior (i.e., Fig. 3(f)) effectively suppresses their presence to a certain degree. Then, we construct $W_p$ into multi-granularity patch-tensor forms to obtain $\mathcal{W}_p$.

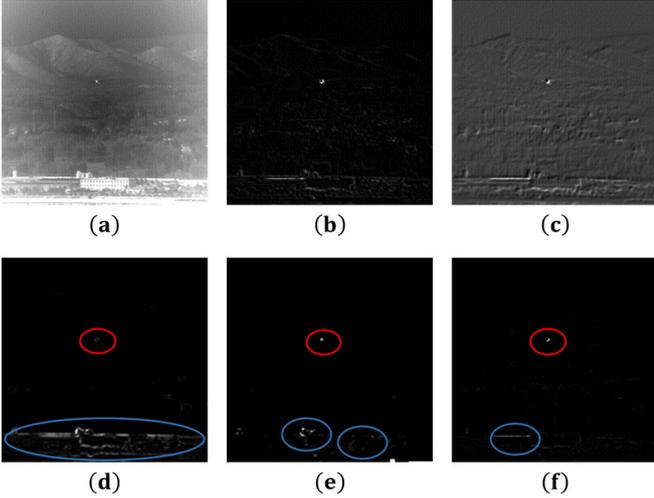

Fig. 3. Comparison of different local structure prior maps. (a) Complex infrared image with small target. (b) Target prior. (c) Background prior. (d) Prior weight map used in RIPT. (e) Prior weight map used in PSTNN. (f) Prior weight map used in the proposed model.

In order to speed up the convergence of the model, we adopt a reweighted strategy [55], which adds a sparse weight:

$$\mathcal{W}_{sw}^{k+1} = \frac{1}{|\mathcal{T}^k| + \varepsilon} \quad (16)$$

where $\varepsilon > 0$ is a small number to prevent the denominator from being 0, and $k + 1$ denotes the $(k + 1)$-th iteration. The local structure tensor is:

$$\mathcal{W} = \mathcal{W}_p' \odot \mathcal{W}_{sw} \quad (17)$$

where $\mathcal{W}_p'$ is the tensor corresponding to the reciprocal of the corresponding element in $\mathcal{W}_p$.

*C. Infrared Target Detection via DWMGIPT Model*

For the MGIPT model, it is generally considered that the background component is slowly transitional and multiple local and nonlocal patches of an image are always highly correlated. Therefore, the background tensor is assumed to be low rank, which is described as follows:

$$\min_{\mathcal{B}} \ \text{rank}_{tt}(\mathcal{B}) \quad (18)$$

Minimizing $\text{rank}_{tt}(\mathcal{B})$ is a NP-hard problem, so we replace the TT rank minimization with the TT nuclear norm, which is a convex surrogate of the former one. The Eq. (18) can be expressed as:

$$\min_{\mathcal{B}} \ \sum_{i=1}^{N-1} \alpha_i \|B_{[i]}\|_* \quad (19)$$

Obviously, the target component $\mathcal{T}$ is a sparse tensor because it contains only a small amount of information compared with the background tensor and can be regularized by $l_1$-norm. For noise component $\mathcal{N}$, it could be considered as additive white Gaussian noise, thus having $\|\mathcal{D} - \mathcal{B} - \mathcal{T}\|_F \leq \delta$ for some $\delta > 0$. Finally, based on the above analysis, we obtain the Tensor Robust Principal Component Analysis (TRPCA) problem which attempts to separate the low-rank and sparse tensors:

$$\min_{\mathcal{B},\mathcal{T}} \ \sum_{i=1}^{N-1} \alpha_i \|B_{[i]}\|_* + \lambda \|\mathcal{T}\|_1$$

$$\text{s.t.} \ \mathcal{D} = \mathcal{B} + \mathcal{T}, \ \|\mathcal{D} - \mathcal{B} - \mathcal{T}\|_F \leq \delta \quad (20)$$

where $\lambda$ is a compromising parameter that controls the tradeoff between the low-rank component and the sparse component.

By integrating the auto-weighted mechanism and the SK-based local structure tensor into the Eq. (20), we propose a DWMGIPT model as follows:

$$\min_{\mathcal{B},\mathcal{T}} \ \sum_{i=1}^{N-1} \alpha_i \|B_{[i]}\|_* + \lambda \|\mathcal{W} \odot \mathcal{T}\|_1$$

$$\text{s.t.} \ \mathcal{D} = \mathcal{B} + \mathcal{T}, \ \|\mathcal{D} - \mathcal{B} - \mathcal{T}\|_F \leq \delta, \ \boldsymbol{\alpha}^T \mathbf{1} = 1, \ \boldsymbol{\alpha} \geq 0$$

$$(21)$$

> REPLACE THIS LINE WITH YOUR MANUSCRIPT ID NUMBER (DOUBLE-CLICK HERE TO EDIT) <*D. Optimization Algorithm*

The optimization problem Eq. (21) can be divided into two blocks. The first block to update $\alpha_i^k$ can be solved by Eq. (6). The second block solving $\mathcal{B}$ and $\mathcal{T}$ can be efficiently solved by ADMM. By introducing additional auxiliary variables $X_i$ and $\mathcal{Y}$, we obtain the following optimization problem:

$$\min_{\mathcal{B},\mathcal{T}} \sum_{i=1}^{N-1} \alpha_i \|X_i\|_* + \lambda \|\mathcal{W} \odot \mathcal{Y}\|_1$$

s.t. $\mathcal{D} = \mathcal{B} + \mathcal{T}$, $\|\mathcal{D} - \mathcal{B} - \mathcal{T}\|_F \leq \delta$, $X_i = B_{[i]}$, $\mathcal{Y} = \mathcal{T}$ (22)

The corresponding augmented Lagrangian function of Eq. (22) is expressed as follows:

$$L = \sum_{i=1}^{N-1} \left\{ \alpha_i \|X_i\|_* + \langle C_i, X_i - B_{[i]} \rangle + \frac{\beta_i}{2} \|X_i - B_{[i]}\|_F^2 \right\}$$

$$+ \lambda \|\mathcal{W} \odot \mathcal{Y}\|_1 + \langle \mathcal{J}, \mathcal{Y} - \mathcal{T} \rangle + \frac{z_1}{2} \|\mathcal{Y} - \mathcal{T}\|_F^2$$

$$+ \langle \mathcal{M}, \mathcal{D} - \mathcal{B} - \mathcal{T} \rangle + \frac{z_2}{2} \|\mathcal{D} - \mathcal{B} - \mathcal{T}\|_F^2 \quad (23)$$

where $C_i$, $\mathcal{J}$, $\mathcal{M}$ represent the Lagrangian multiplier and $\beta_i$, $z_1$, $z_2$ are positive penalty scalars. Eq. (23) is decomposed into 3 optimization sub-problems through ADMM, including $X_i$, $\mathcal{Y}$ and ($\mathcal{B}$, $\mathcal{T}$). The solution details are given as follows:

*1) Updating $X_i$ with Other Variables Fixed:*

$$X_i^{k+1} = \underset{X_i}{\operatorname{argmin}} \sum_{i=1}^{N-1} \left\{ \alpha_i^k \|X_i\|_* + \langle C_i^k, X_i - B_{[i]}^k \rangle + \frac{\beta_i}{2} \|X_i - B_{[i]}^k\|_F^2 \right\} = \underset{X_i}{\operatorname{argmin}} \sum_{i=1}^{N-1} \left\{ \alpha_i^k \|X_i\|_* + \frac{\beta_i}{2} \left\| X_i - B_{[i]}^k + \frac{C_i^k}{\beta_i} \right\|_F^2 \right\} \quad (24)$$

The above optimization problem's closed-form solution is:

$$X_i^{k+1} = \operatorname{SVT}_{\tau_i^k}\left(B_{[i]}^k - \frac{C_i^k}{\beta_i}\right) = U \operatorname{diag}\left(\max(s_{r,r} - \tau_i^k, 0)\right) V^T \quad (25)$$

where $B_{[i]}^k - \frac{C_i^k}{\beta_i} = USV^T$, $s_{r,r}$ is the $r$-th singular value of $S$, $\tau_i^k = \alpha_i^k / \beta_i$ and $\operatorname{SVT}_{\tau_i^k}(\cdot)$ denotes the singular value thresholding (SVT) operator [56].

*2) Updating $\mathcal{Y}$ with Other Variables Fixed:*

$$\mathcal{Y}^{k+1} = \underset{\mathcal{Y}}{\operatorname{argmin}} \lambda \|\mathcal{W} \odot \mathcal{Y}\|_1 + \langle \mathcal{J}^k, \mathcal{Y} - \mathcal{T}^k \rangle + \frac{z_1^k}{2} \|\mathcal{Y} - \mathcal{T}^k\|_F^2 = \underset{\mathcal{Y}}{\operatorname{argmin}} \lambda \|\mathcal{W} \odot \mathcal{Y}\|_1 + \frac{z_1^k}{2} \left\| \mathcal{Y} - \mathcal{T}^k + \frac{\mathcal{J}^k}{z_1^k} \right\|_F^2 \quad (26)$$

The above optimization problem can be solved by the soft shrinkage operator [57]:

$$\mathcal{Y}^{k+1} = \operatorname{Th}_{\frac{\lambda \mathcal{W}}{z_1^k}}\left( \mathcal{T}^k - \frac{\mathcal{J}^k}{z_1^k} \right) \quad (27)$$

where $\operatorname{Th}_\lambda(x) = \operatorname{sign}(x) \cdot \max(|x| - \lambda, 0)$.

*3) Updating $\mathcal{B}$ and $\mathcal{T}$ with Other Variables Fixed:*

$$(\mathcal{B}^{k+1}, \mathcal{T}^{k+1}) = \underset{\mathcal{B},\mathcal{T}}{\operatorname{argmin}} \sum_{i=1}^{N-1} \frac{\beta_i}{2} \left\| X_i^{k+1} - B_{[i]} + \frac{C_i^k}{\beta_i} \right\|_F^2 + \frac{z_1^k}{2} \left\| \mathcal{Y}^{k+1} - \mathcal{T} + \frac{\mathcal{J}^k}{z_1^k} \right\|_F^2 + \frac{z_2^k}{2} \left\| \mathcal{D} - \mathcal{B} - \mathcal{T} + \frac{\mathcal{M}^k}{z_2^k} \right\|_F^2 \quad (28)$$

For the least squares problem Eq. (28), by taking the derivative of $\mathcal{B}$ and $\mathcal{T}$, respectively, we have:

$$(\sum_{i=1}^{N-1} \beta_i + z_2^k) \mathcal{B} + z_2^k \mathcal{T} = \sum_{i=1}^{N-1} \beta_i \left( \operatorname{unreshape}_{[i]}\left(X_i^{k+1} + \frac{C_i^k}{\beta_i}\right) \right) + z_2^k \left( \mathcal{D} + \frac{\mathcal{M}^k}{z_2^k} \right)$$

$$z_2^k \mathcal{B} + (z_1^k + z_2^k) \mathcal{T} = z_1^k \left( \mathcal{Y}^{k+1} + \frac{\mathcal{J}^k}{z_1^k} \right) + z_2^k \left( \mathcal{D} + \frac{\mathcal{M}^k}{z_2^k} \right) \quad (29)$$

Then the $\mathcal{B}$ and $\mathcal{T}$ can be precisely obtained as follows:

$$\mathcal{B}^{k+1} = \frac{\left(z_2^k \mathcal{H}^k - (z_1^k + z_2^k) \mathcal{G}^k\right)}{z_2^{k^2} - (\sum_{i=1}^{N-1} \beta_i + z_2^k)(z_1^k + z_2^k)} \quad (30)$$

and

$$\mathcal{T}^{k+1} = \frac{\left(z_2^k \mathcal{G}^k - (\sum_{i=1}^{N-1} \beta_i + z_2^k) \mathcal{H}^k\right)}{z_2^{k^2} - (\sum_{i=1}^{N-1} \beta_i + z_2^k)(z_1^k + z_2^k)} \quad (31)$$

where $\mathcal{G}^k = \sum_{i=1}^{N-1} \beta_i \left( \operatorname{unreshape}_{[i]}(X_i^{k+1} + C_i^k / \beta_i) \right) + z_2^k (\mathcal{D} + \mathcal{M}^k / z_2^k)$ and $\mathcal{H}^k = z_1^k (\mathcal{Y}^{k+1} + \mathcal{J}^k / z_1^k) + z_2^k (\mathcal{D} + \mathcal{M}^k / z_2^k)$.

*4) Updating Multipliers $C_i$, $\mathcal{J}$, $\mathcal{M}$ and Penalty Factors $z_1$, $z_2$ with Other Variables Fixed:*

$$C_i^{k+1} = C_i^k + \beta_i (X_i^{k+1} - B_{[i]}^{k+1}) \quad (32)$$

$$\mathcal{J}^{k+1} = \mathcal{J}^k + z_1^k (\mathcal{Y}^{k+1} - \mathcal{T}^{k+1}) \quad (33)$$

$$\mathcal{M}^{k+1} = \mathcal{M}^k + z_2^k (\mathcal{D} - \mathcal{B}^{k+1} - \mathcal{T}^{k+1}) \quad (34)$$

$$z_1^{k+1} = \rho z_1^k \quad (35)$$

$$z_2^{k+1} = \rho z_1^k \quad (36)$$

Algorithm 1 summarizes the detailed calculation process based on ADMM.

*E. Infrared Target Detection Procedure*

The whole procedure of proposed DWMGIPT method can be described as follows:

**Algorithm 1** Optimization Framework
---
**Input:** The observed tensor $\mathcal{D} \in \mathbb{R}^{I_1 \times I_2 \times \cdots \times I_N}$, parameters $\beta, \lambda$
**Initialize:** $\mathcal{B}^0 = \mathcal{D}$, $\mathcal{T}^0 = 0$, $\alpha_i^0 = \frac{1}{N-1}$, $\mathcal{M}^0 = 0$, $\mathcal{J}^0 = 1$, $C_i^0 = 1$, $i = 1, 2, \cdots, N-1$, $z_1 = 0.15$, $z_2 = 0.02$, $\rho = 1.2$, $\zeta = 1e-3$, maximum iteration step $K = 200$.
1: **While** not converged **Do**
2: Update $X_i^{k+1}$ via Eq. (25)
3: Update $\mathcal{Y}^{k+1}$ via Eq. (26)
4: Update $\mathcal{B}^{k+1}$ via Eq. (30)
5: Update $\mathcal{T}^{k+1}$ via Eq. (31)
6: Update $C_i^{k+1}$, $\mathcal{J}^{k+1}$, $\mathcal{M}^{k+1}$, $z_1^{k+1}$, $z_2^{k+1}$ via Eq. (32) - Eq. (36)
7: Update $\alpha_i^{k+1}$ via Eq. (6)
8: Update $\mathcal{W}^{k+1}$ via Eq. (17)
9: Check the convergence condition
$\frac{\left\|\mathcal{D} - \mathcal{B}^{k+1} - \mathcal{T}^{k+1}\right\|_F^2}{\|\mathcal{D}\|_F^2} \leq \zeta$
10: Update $k = k + 1$
11: **End While**
**Output:** Background component $\mathcal{B}$ and target component $\mathcal{T}$.

1) Local Prior Extraction. Given an infrared image, through the steering kernel theory, the prior weight map $W_p$, which integrates target information and background information, is obtained.
2) Multi-granularity patch-tensor construction. By using a sliding window of size $m \times n$ from top left to bottom right to traverse the original infrared image and utilizing prime factorization to achieve tensor augmentation, the multi-granularity patch-tensor $\mathcal{D} \in \mathbb{R}^{m_1 \times \cdots \times m_i \times n_1 \times \cdots \times n_j \times cols \times rows}$ corresponding to the original infrared image can be constructed.
3) Target and background separation. The original tensor $\mathcal{D}$ is decomposed into target tensor $\mathcal{T}$ and background tensor $\mathcal{B}$ through Algorithm 1.
4) Image reconstruction. The target image $f_T$ and background image $f_B$ can be reconstructed by the inverse operation of construction. Once the reconstruction is completed, small targets are extracted through adaptive threshold segmentation approach in [28].

*F. Convergence Analysis*

In this subsection, we briefly analyze and verify the convergence of the proposed DWMGIPT model. The ADMM algorithm is widely used to solve convex optimization problem, and the theoretical analysis of its convergence can be found in [58] and [59]. According to the Weierstrass theorem [58], the objective function Eq. (22) is continuous, convex, and nonempty. Then, the convergence of Algorithm 1 can be obtained according to the reference [60], which presents a detailed proof. In addition, in our algorithm, we use an empirical convergence condition $\frac{\left\|\mathcal{D} - \mathcal{B}^{k+1} - \mathcal{T}^{k+1}\right\|_F^2}{\|\mathcal{D}\|_F^2} \leq \zeta$ to verify the convergence. Fig. 4 shows the variation of the objective function value on Seq.1. As observed in result, the objective function converges to zero when $k \geq 40$.

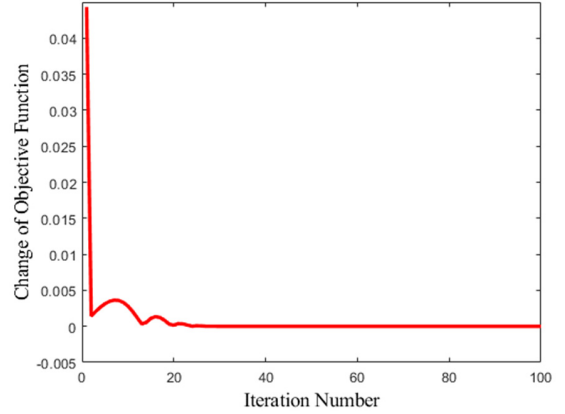

Fig. 4. The variation of the objective function obtained by DWMGIPT on Seq.1.

IV. EXPERIMENT

In this section, we first introduce test datasets and quantitative evaluation metrics for infrared small target detection. Next, we discuss the impact of different parameter settings on the DWMGIPT. Finally, in order to fully verify the effectiveness of the proposed algorithm, extensive experiments are implemented to compare it with eight state-of-the-art approaches in various complex scenes.

*A. Data Preparation*

The test data used in the experiment includes 20 infrared image sequences with a total of 1500 images, including the public real data [61,62] and 5 infrared image sequences simulated by the approach in [28]. They include not only various challenging scenes such as building, mountain, vegetation, sea, and sky taken from different perspectives, but also interference clutter such as cloud, noise, and various ground highlight areas. Meanwhile, all the infrared targets are of small size, lacking texture structure and color features. With the aim of enhancing the target clarity, these images are converted to a uniform size. As shown in Fig. 5, the small target is marked by a red rectangle whose area is also enlarged. Fig. 5(a)-(t) are representative frames selected from 20 infrared sequences, from which we can see that in most scenes the targets are tiny and the backgrounds are complicated and volatile. TABLE I shows the target and scene descriptions of the test data. In order to verify the robustness of the algorithm for different scenes, we classify the 15 public real infrared image sequences into Data1, Data2 and Data3 according to three scenes of buildings, mountains and vegetation. In addition, all experiments and simulations are implemented with MATLAB R2022b in Windows 10 based on AMD Ryzen 7 5800H 3.20 GHz CPU with 16GB memory.

TABLE I
DESCRIPTION OF TEST DATA

| Test data | Frames | Image Size | Description of target | Description of background |
|---|---|---|---|---|
| Data.1 | 300 | 256×256 640×512 | Tiny, varying target intensity | Low-altitude urban and rural background with various reflections |
| Data.2 | 300 | 256×256 | Tiny, varying size, low target intensity | Mountains with various reflections |
| Data.3 | 300 | 256×256 640×512 | Tiny, varying size, low target intensity | Vegetation and reflective road |
| Seq.1 | 120 | 256×256 | Tiny, low target intensity, fast movement | Gloomy mountain background with strong noise |
| Seq.2 | 120 | 256×256 | Tiny, moving along cloud edges | Multilayer cloud and strong noise |
| Seq.3 | 120 | 256×256 | Tiny, moving over sea | Gloomy sea and land background with strong noise |
| Seq.4 | 120 | 256×256 | Tiny, low target intensity | Cloud, building, field and slight noise |
| Seq.5 | 120 | 256×205 | Tiny, moving through clouds | Thick cloud and slight noise |

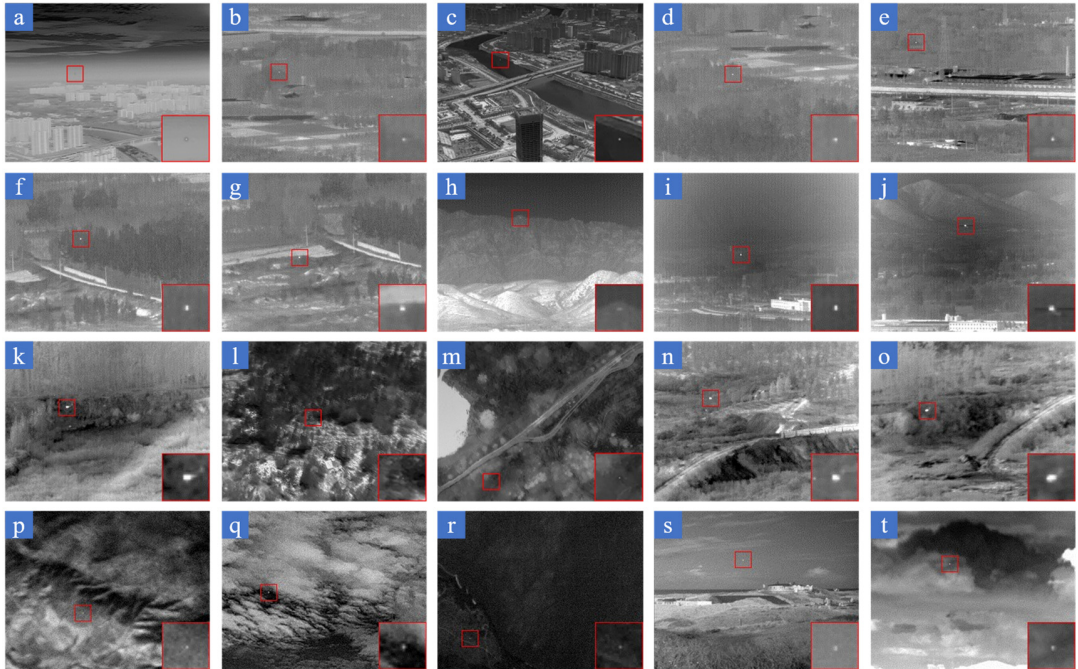

Fig. 5. (a)-(e), (f)-(j), (k)-(o) and (p)-(t) are representative frames selected from Data1, Data2, Data3 and Seq1-5, respectively.

*B. Evaluation Metrics*

We adopt commonly evaluation metrics, including SCR gain (SCRG), background suppression factor (BSF), and receiver operating characteristic curve (ROC) to quantitatively validate the performance of the infrared small target detection algorithms. The ROC curve is a comprehensive representation that reveals the trade-off between the algorithm's detection probability $P_d$ and false-alarm rate $F_a$. Taking $P_d$ as the y-axis and $F_a$ as the x-axis, we can construct the ROC curve and assess the area under the curve (AUC).

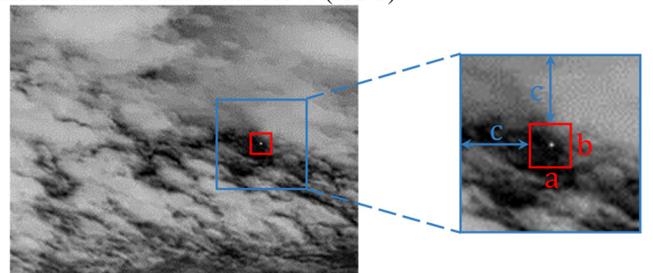

Fig. 6. Schematic diagram of the target neighborhood.

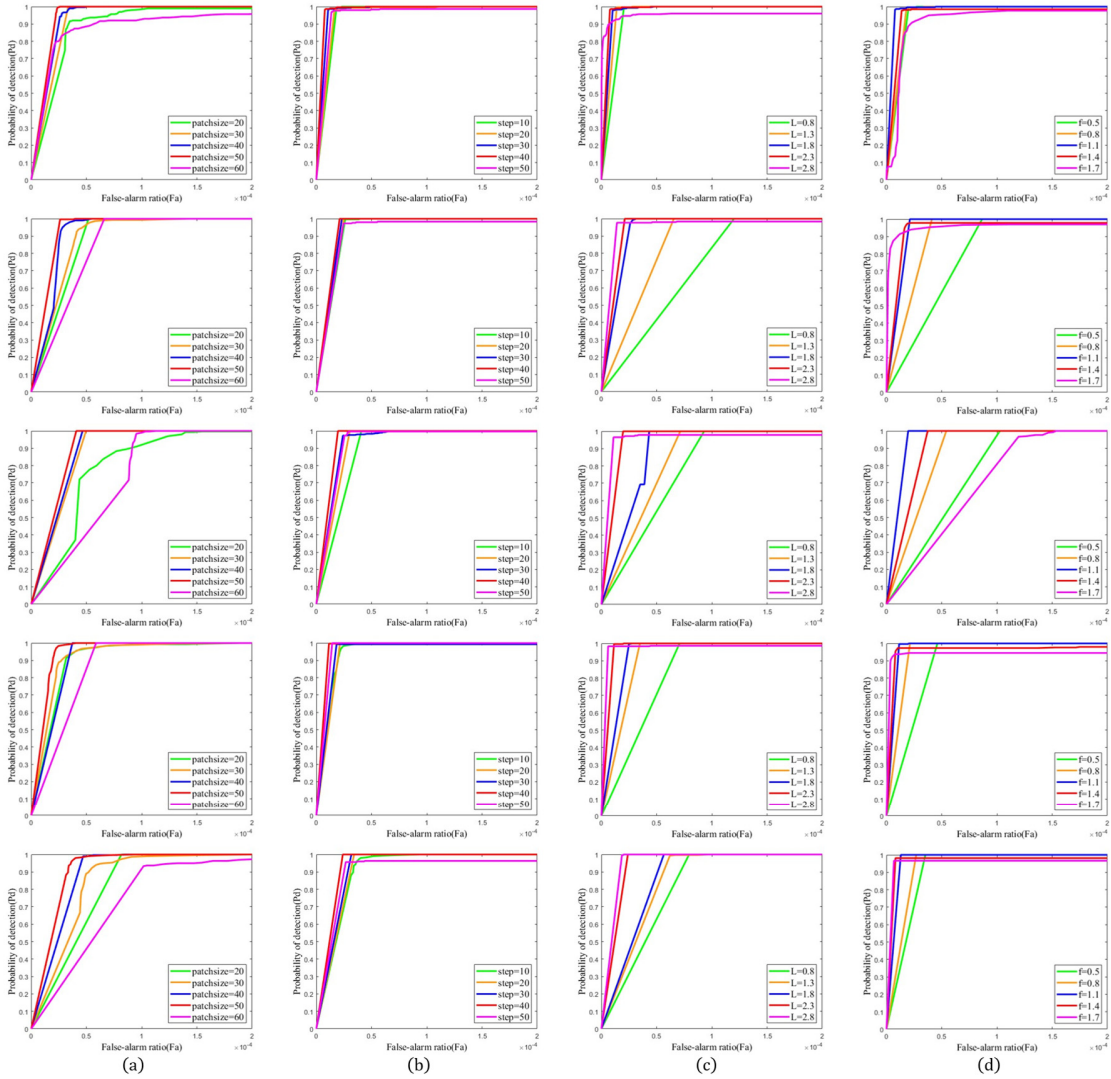

Fig. 7. ROC curves of different parameters in the test data. (a) Patch Size. (b) Sliding Step Size. (c) Compromising Factor. (d) Penalty Factor.

Generally, the larger the AUC value, the better the detection performance. The calculation formula is defined as follows [39]:

$$P_d = \frac{\text{number of true detections}}{\text{number of actual targets}} \quad (37)$$

$$F_a = \frac{\text{number of false detections}}{\text{number of image pixels}} \quad (38)$$

Besides, BSF is used to compare the background suppression performance of all algorithms, which is expressed as:

$$\text{BSF} = \frac{\sigma_{\text{in}}}{\sigma_{\text{out}}} \quad (39)$$

where $\sigma_{\text{in}}$ and $\sigma_{\text{out}}$ represent the standard deviation of the target neighborhood in the original image and the processed image, respectively. The SCRG reflects the enhancement effect of the target by calculating the signal-to-clutter ratio (SCR) before and after processing, which is defined as:

$$\text{SCRG} = \frac{\text{SCR}_{\text{out}}}{\text{SCR}_{\text{in}}} \quad (40)$$

where SCR reflects the degree of discrimination between the target and the background clutter and can be utilized to assess the difficulty of detecting infrared small targets. The calculation formula of SCR is as follows:

$$\text{SCR} = \frac{|\mu_t - \mu_b|}{\sigma_b} \quad (41)$$

where $\mu_t$ represents the average value of the target area, $\mu_b$ and $\sigma_b$ represent the average value and the standard deviation of the target neighborhood. Considering the possibility of the standard deviation being zero after background suppression, we refer to [63] to incorporate the adjustment coefficient $\phi$ into the calculations of SCR and BSF. In this paper, $\phi$ is empirically set to 0.01. The target area and the target neighborhood are shown in Fig. 6. The target area size is $a \times b$ and the target neighborhood size is $(a + 2c) \times (b + 2c)$. In this paper, we follow [39]to set $c = 65$ in the experiment.

TABLE II
PARAMETER OF NINE METHODS

| Methods | Parameters |
| --- | --- |
| Top-hat | Shape: disk, size: 5×5. |
| TLLCM | Different filtering window: 3×3, 5×5, 7×7. |
| IPI | Patch size: 50×50, step: 10, $\lambda = 1/\sqrt{min(m,n)}$, $\varepsilon = 10^{-7}$. |
| NRAM | Patch size: 50×50, step: 10, $\lambda = 1/\sqrt{min(m,n)}$, $\varepsilon = 10^{-7}$, $\mu^0 = 3\sqrt{min(m,n)}$, $\gamma = 0.002$, $C = \sqrt{min(m,n)}/2.5$. |
| RIPT | Patch size: 50×50, step: 10, $\lambda = 1/\sqrt{min(m,n)}$, $\varepsilon = 10^{-7}$, $h = 1$. |
| PSTNN | Patch size: 40×40, step: 40, $\lambda = 0.7/\sqrt{min(n_1, n_2) * n_3}$, $\varepsilon = 10^{-7}$. |
| NTFRA | Patch size: 40×40, step: 40, $\lambda = 1/\sqrt{min(n_1, n_2) * n_3}$, $\beta = 0.05$, $\mu = 200$. |
| ANLPT | Patch size: 50×50, step: 50, region: 10, channel: 3, $\mu = 10^{-3}$. |
| Proposed | Patch size: 50×50, step: 40, $\lambda = 2.3/\sqrt{ps_1 * \cdots * ps_n * z}$, $f = 1.1$, $z_1 = 0.15$, $z_2 = 0.02$. |

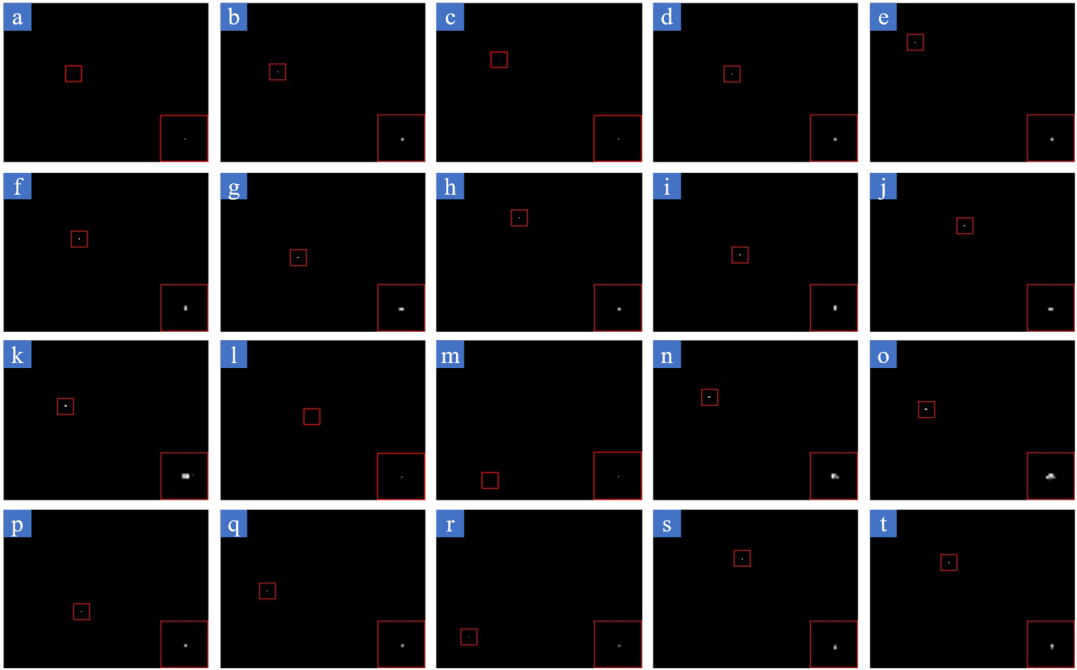

Fig. 8. Detection result of (a)-(t) 20 scenes, respectively.

*C. Parameter Analysis*

In this subsection, in order to make the algorithm have better performance, we briefly analyze the key parameters of the DWMGIPT model including the patch size, the step size of the sliding window, the compromising factor $\lambda$, and the penalty factor $\beta_i$. As shown in Fig. 7, parameters are selected within a certain range based on previous experience, and the optimal parameters are determined by evaluating the ROC curve on Data.1-3 and Seq.1-2. Due to the limitation of verifying the most optimal value for all parameters simultaneously, a method is adopted where each parameter is adjusted while keeping the remaining parameters fixed. This process enables the determination of the local optimal value of the model corresponding to each parameter.

*1) Patch Size:* The size of the infrared image patch not only affects the detection performance, but also determines the computational complexity of the algorithm. In order to guarantee the sparsity of the target, we hope for a larger patch size. However, at the same time, interference clutter such as system noise and strong edges may also be recognized as target, which reduces the detection performance. On the other hand, if the patch size is too small, the constructed multi-granularity patch tensor contains insufficient information, leading to the disappearance of the relationship between the target and the background. We change the patch size from 20 to 60 with ten intervals and the corresponding ROC curves are shown in Fig. 7(a). The analysis of the ROC curves demonstrates that the algorithm achieves its optimal performance when the patch size is set to 50.

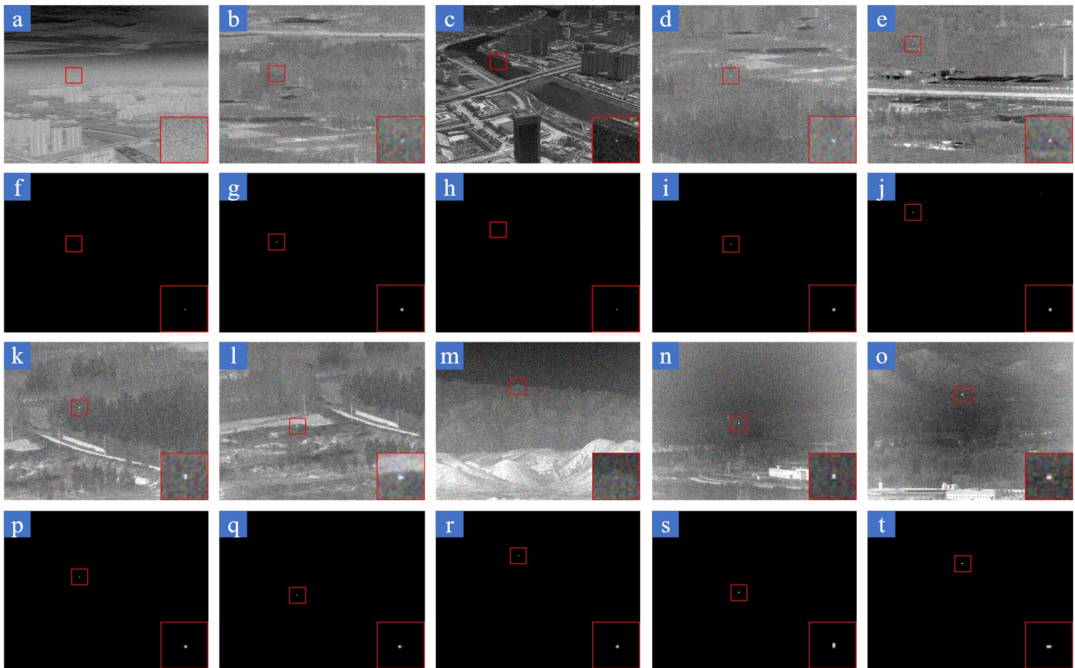

Fig. 9. Original images after adding Gaussian noise with $\sigma = 15$ and detection results.

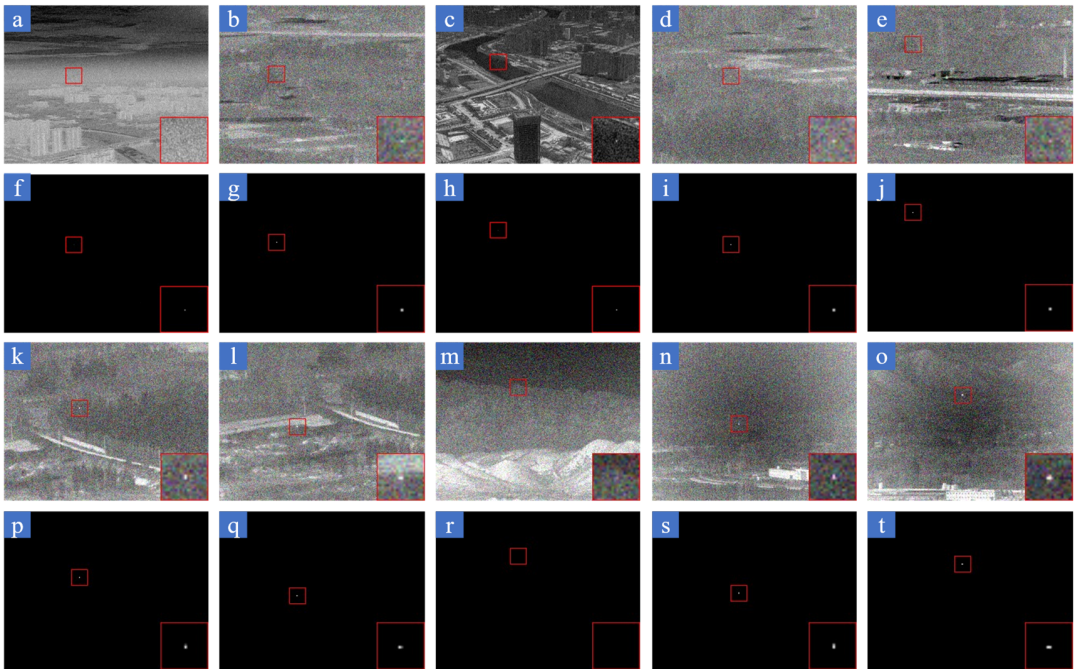

Fig. 10. Original images after adding Gaussian noise with $\sigma = 25$ and detection results.

*2) Sliding Step Size:* Similar to the patch size, the detection performance and calculation time can be influenced by the choice of sliding step size. When the sliding step size is large, the acquired image patches will be reduced, which disrupts the nonlocal self-correlation of the background. When the sliding step size is too small, it will increase the time cost of SVD and constructing the multi-granularity patch tensor. In order to investigate actual influence of the sliding step size, we change it from 10 to 50 with ten intervals and the corresponding ROC curves are shown in Fig. 7(b). Considering algorithm complexity and ROC curve, the model achieves its best performance when sliding step size is set to 40.

*3) Compromising Factor $\lambda$ :* The compromising factor $\lambda$ controls the balance between the sparse target and the low-rank background in the model. If $\lambda$ is too large, the target will shrink too much, potentially leading to the loss of necessary information. If $\lambda$ is too small, the background clutter cannot be completely suppressed, resulting in many false alarms. In the experiment, we set $\lambda = L / (ps_1 \times \cdots \times ps_n \times z)^{1/2}$ and $ps_1, \cdots, ps_n$ are the prime factors of patch size. Therefore, if the input tensor is known, $\lambda$ depends on the size of L. We change L from 0.8 to 2.8 with an interval of 0.5 and the corresponding ROC curves are shown in Fig. 7(c). Considering both the detection probability and false-alarm rate comprehensively, we set the optimal value of L to 2.3 in the following experiment.

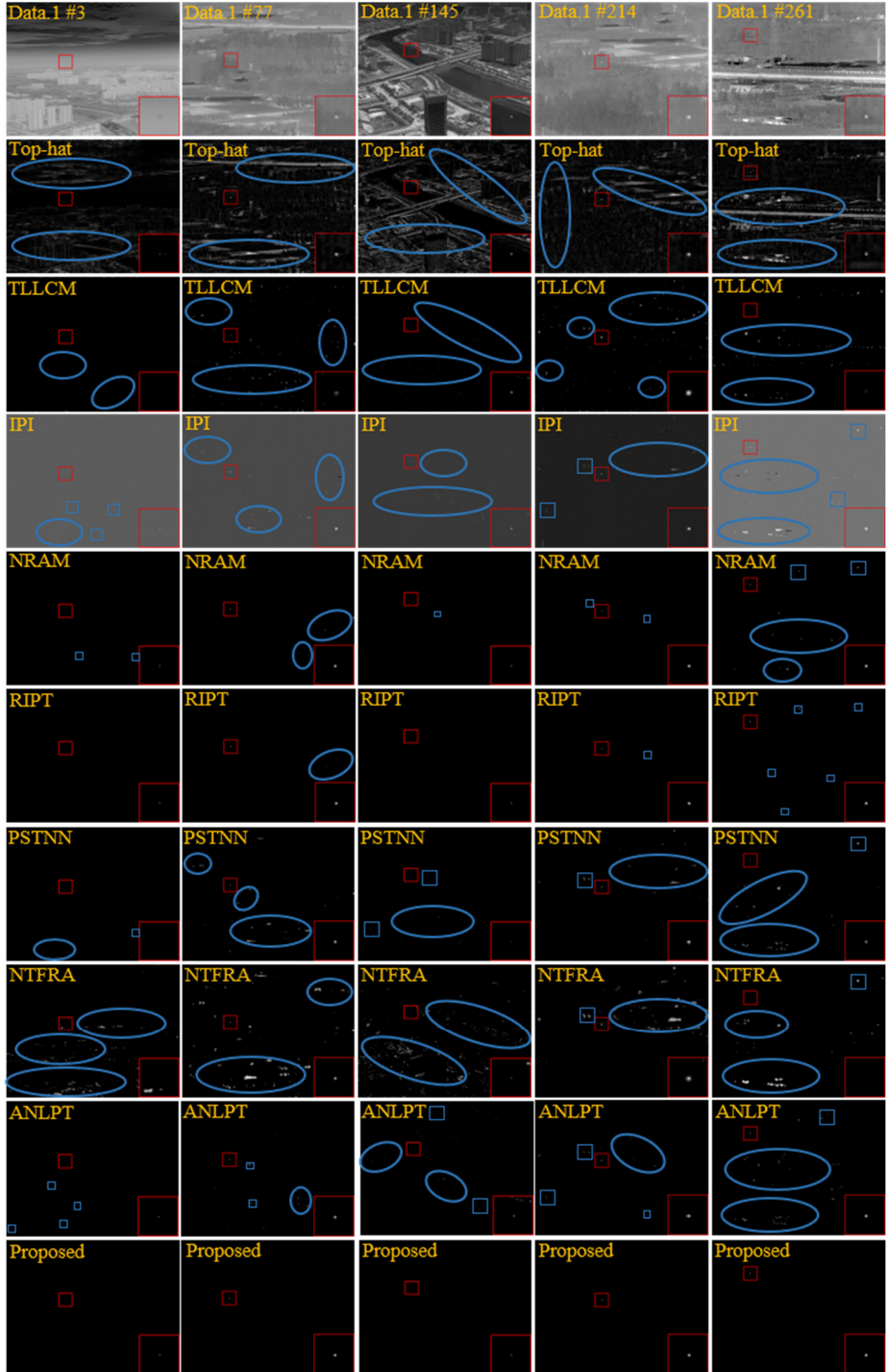

Fig. 11. Comparative results of different methods on Data.1. The blue rectangles and ellipses denote background and noise residues. For enhancing visual clarity, the target area marked by the red rectangle is zoomed in the right bottom corner.

*4) Penalty Factor $\beta_i$:* The penalty factor $\beta_i$ controls the singular value thresholding operator. When $\beta_i$ is too small, noise and clutter cannot be well suppressed. As the value of $\beta_i$ increases, the performance of the algorithm to suppress background clutter gradually improves. However, if $\beta_i$ is too large, it will smooth out the target as clutter and even cause the model to terminate the iteration in advance and damage the detection performance. Since $\beta_i$ plays an important role in detection performance and model convergence speed, it must be chosen carefully. According to experience, $\beta_i$ can be expressed as $\beta_i = f \times \alpha_i$. In case of giving the $\alpha_i$, $\beta_i$ depends on the value of f. We change f from 0.5 to 1.7 with an interval of 0.3 and the corresponding ROC curves are shown in Fig. 7(d). The ROC curve reveals that the value of f should be set to 1.1 for better performance.

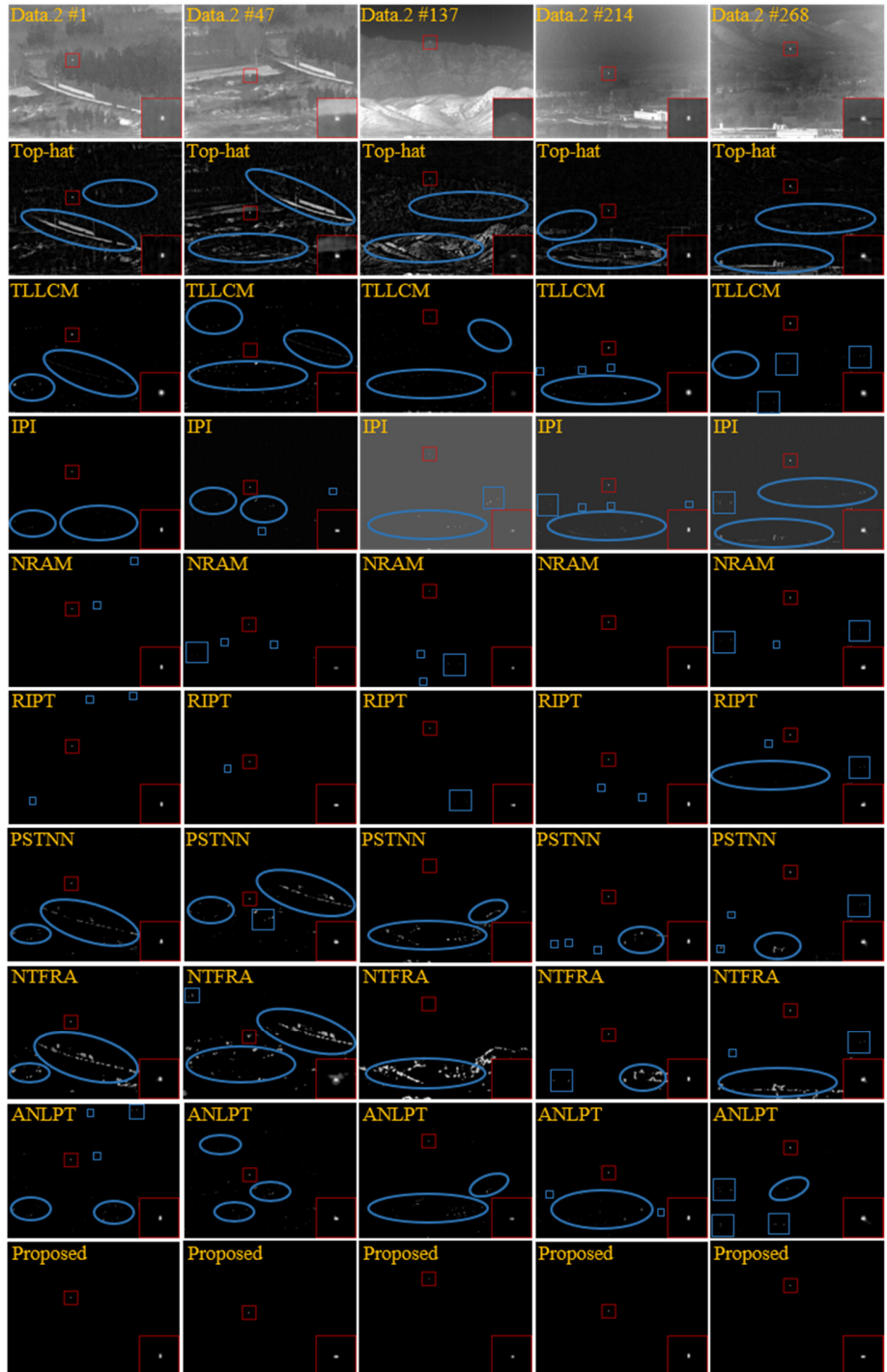

Fig. 12. Comparative results of different methods on Data.2. The blue rectangles and ellipses denote background and noise residues. For enhancing visual clarity, the target area marked by the red rectangle is zoomed in the right bottom corner.

*D. Experiment Results*

In the experiment, we compare the proposed algorithm with eight state-of-the-art approaches: Top-hat [15], TLLCM [27], IPI [28], NRAM [34], RIPT [37], PSTNN [38], NTFRA [39], ANLPT [40], and the detailed parameter settings of those are shown in TABLE II.

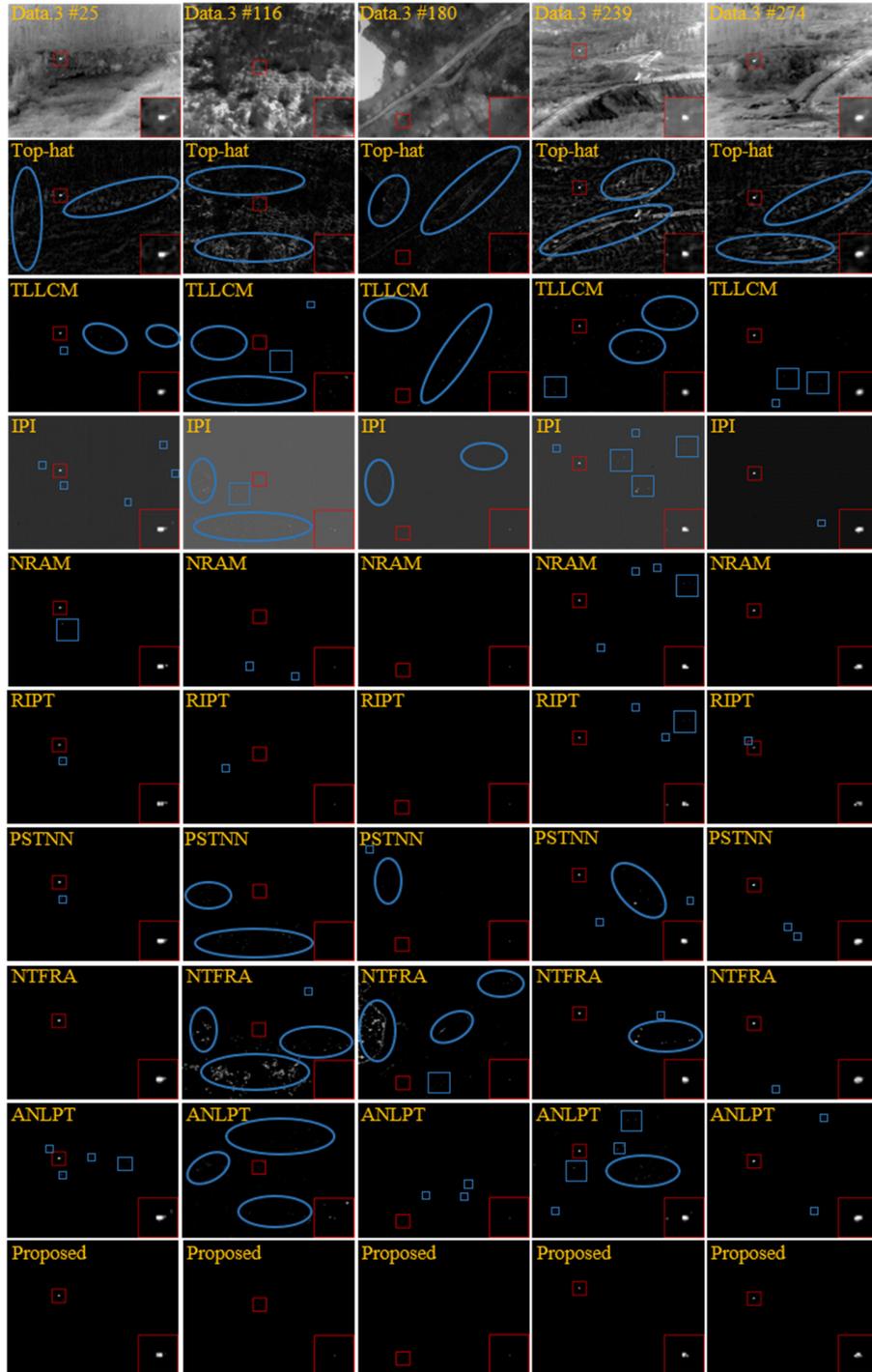

Fig. 13. Comparative results of different methods on Data.3. The blue rectangles and ellipses denote background and noise residues. For enhancing visual clarity, the target area marked by the red rectangle is zoomed in the right bottom corner.

*1) Scene Robustness:* In order to verify the robustness of our algorithm in different scenes, we randomly select a frame as the representative image from 20 different sequences, as shown in Fig. 5. The corresponding detection results in Fig. 8 demonstrate that our algorithm can detect the target relatively completely and suppress the background clutter successfully. However, in scenes (k), scenes (n) and scenes (o), because of the irregular shape of some big targets, the shape of the detected target may be incomplete. In general, our method can detect small targets and suppress clutter interference in various challenging scenes.

*2) Noise Robustness:* In real scenes, noise is also a crucial factor affecting the detection results. Therefore, we evaluate the performance of the proposed model in different scenes with different levels of noise. To verify its robustness to noise, we added Gaussian noise of σ = 15 and σ = 25 to ten challenging scenes. As can be seen from Fig. 9 (a)-(e) and (k)-(o), when the standard deviation is 15, the noise scatters across the entire image randomly dispersed, blurring the contour of the small target, and even in (m) the small target is almost overwhelmed in the background and noise. Fortunately, as can be seen from Fig. 9 (f)-(j) and (p)-(t), our method can obtain good detection results in heavy noisy scenes. As shown in Fig. 10, when the noise standard deviation increases to 25, the contour and intensity of the target weaken further, the proposed model can still detect the target successfully, but fails in scene (m). However, this failure can be considered acceptable due to the target is completely overwhelmed in the noise. In summary, as long as the target maintains a certain contrast in the polluted image, the proposed method is able to enhance target and suppress noise well.

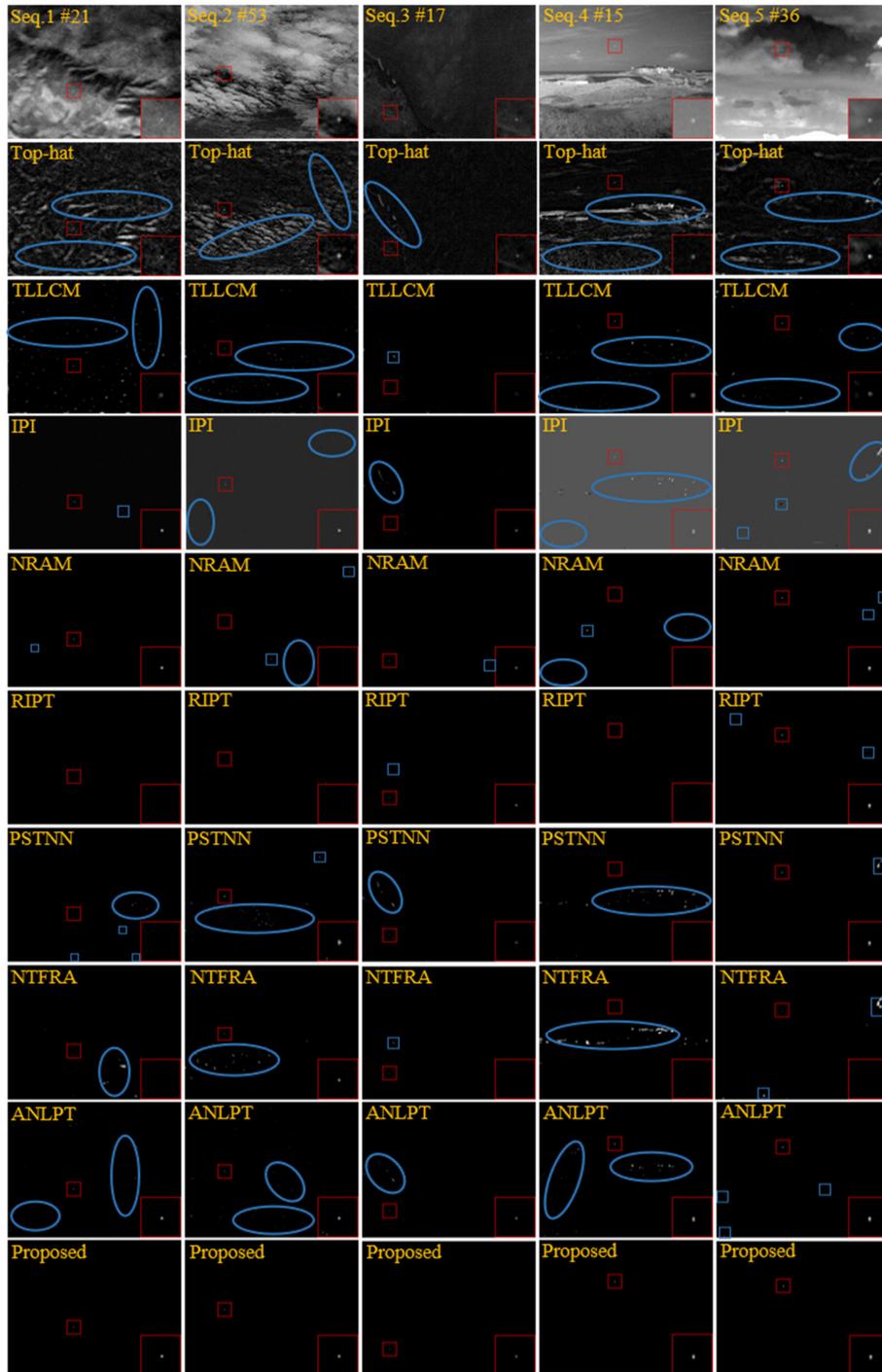

Fig. 14. Comparative results of different methods on Seq.1-5. The blue rectangles and ellipses denote background and noise residues. For enhancing visual clarity, the target area marked by the red rectangle is zoomed in the right bottom corner.

*3) Visual Comparison:* In the twenty different infrared sequences, the proposed algorithm is compared with other eight state-of-the-art models, and one frame is randomly selected from each sequence as representative results. The comparative results of Data.1-3 are shown in Fig. 11-13 and the detection results of Seq.1-5 are shown in Fig. 14. There are lots of clutter residuals of reflective areas and noise in the detection results of the Top-hat. Compared with Top-hat, TLLCM is significantly improved in suppressing interference clutter but sometimes cannot detect the target as shown in Fig. 11. The background of the IPI detection results is gray and there are many background residuals, indicating that the method basically cannot completely suppress the background. Although NRAM, RIPT and ANLPT performs better in suppressing clutter interference. In Seq.2 and Seq.4, NRAM and RIPT miss the detection of targets, and RIPT even has the undetectable result. PSTNN and NTFRA can detect simple scenes with relatively large target perfectly in Data.3, but fail to separate small target and background effectively in complex scenes such as Data.1, Data.2 and Seq.1-5, and many clutter pixels still remain in the detection result. On the contrast, the proposed algorithm not only can completely suppress background interference and noise but also extract the target shape relatively accurately. It is worth noting that the data set includes twenty complex scenes, so the experimental results can prove the robustness and superiority of the proposed algorithm.

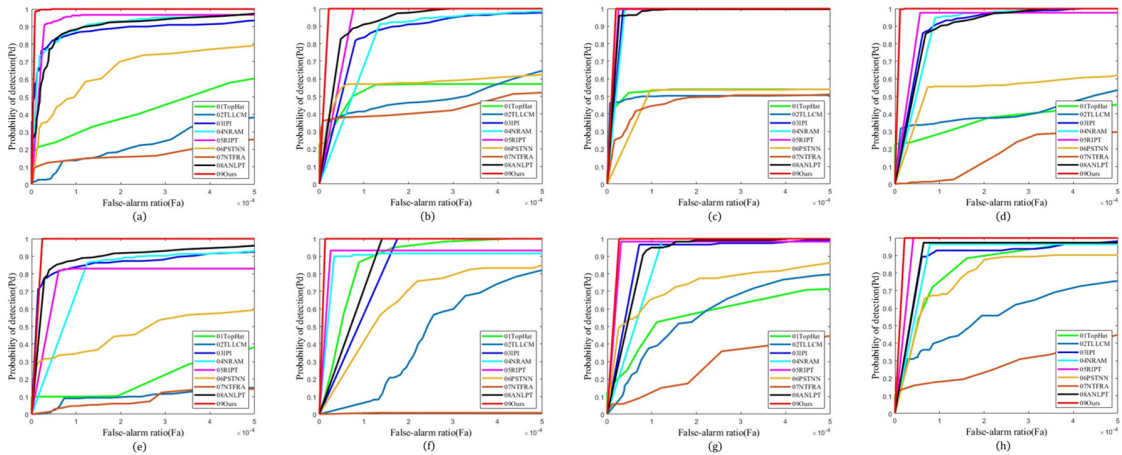

Fig. 15. ROC curves of nine methods on test data.

TABLE III
AUC OF NINE METHODS

| Test data | Top-hat | TLLCM | IPI | NRAM | RIPT | PSTNN | NTFRA | ANLPT | Proposed |
|---|---|---|---|---|---|---|---|---|---|
| Data.1 | 0.4128 | 0.2271 | 0.8772 | 0.9050 | **0.9336** | 0.6517 | 0.1749 | 0.8926 | **0.9917** |
| Data.2 | 0.5327 | 0.4811 | 0.8471 | 0.8208 | **0.9327** | 0.5661 | 0.4280 | 0.9183 | **0.9789** |
| Data.3 | 0.5264 | 0.4970 | 0.9644 | 0.9605 | **0.9712** | 0.4852 | 0.4627 | 0.9651 | **0.9804** |
| Seq.1 | 0.3646 | 0.3997 | 0.9042 | 0.8945 | **0.9216** | 0.5350 | 0.1696 | 0.8944 | **0.9885** |
| Seq.2 | 0.1992 | 0.1022 | 0.8581 | 0.7858 | 0.7786 | 0.4648 | 0.0871 | **0.8870** | **0.9760** |
| Seq.3 | 0.8766 | 0.4316 | 0.8250 | 0.8833 | **0.9097** | 0.6423 | 0.0072 | 0.8595 | **0.9870** |
| Seq.4 | 0.5605 | 0.5617 | 0.9044 | 0.8629 | **0.9527** | 0.7263 | 0.2755 | 0.9007 | **0.9732** |
| Seq.5 | 0.8308 | 0.5560 | 0.8836 | 0.8888 | **0.9588** | 0.7800 | 0.2779 | 0.9109 | **0.9785** |

*4) Quantitative Evaluation:* In this section, ROC, AUC, SCRG and BSF four evaluation metrics are used to compare nine methods quantitatively. The ROC curves of different comparison methods are shown in Fig. 15. Generally, the closer the ROC curve to the upper left corner, the better the detection performance of the corresponding method. As can be seen from Fig. 15 (a)-(h), our method is closer to the upper left corner and achieves higher $P_d$ than other algorithms under the same $F_a$. In order to compare the detection performance of the algorithm quantitatively, TABLE III shows the AUC values in different test data, where the maximum and second maximum values of AUC are represented by red font and green font, respectively. The AUC values of the proposed algorithm are higher than other methods and closest to 1 on all test data, which indicating that ours has better detection performance.

The SCRG and BSF of the nine methods listed in TABLE IV are used to reflect the target enhancement ability and the background suppression ability. The red font and green font denote the maximum and second maximum values of SCRG and BSF in each test data, respectively. It should be noted that we do not show the indicator results of the RIPT method on the Seq.1, Seq.2 and Seq.4. The main reason is that the method obtains undetectable results in these scenes, it becomes meaningless for calculating indicator value. As can be observed from TABLE IV, NRAM and RIPT can effectively suppress clutter interference on some test data such as Data.1 and Seq.5, but their performance decreases when applied to other test data. Furthermore, PSTNN based on TNN and NTFRA improved on the basis of TNN cannot suppress clutter interference and even lose target information when dealing with small targets with complex backgrounds, because they lack the ability to estimate low-rankness of the background from multiple modes. The indicator values in the Table shows that Our algorithm achieves better target enhancement and background suppression performance.

TABLE IV
SCRG AND BSF IN NINE METHODS

| Test data | Metrics | Top-hat | TLLCM | IPI | NRAM | RIPT | PSTNN | NTFRA | ANLPT | Proposed |
|---|---|---|---|---|---|---|---|---|---|---|
| Data.1 | SCRG | 3.93 | 12.67 | 18.06 | 37.94 | **48.08** | 16.72 | 5.84 | 26.11 | **56.72** |
|  | BSF | 0.99 | 3.32 | 4.58 | 6.72 | **7.26** | 4.04 | 2.46 | 4.37 | **7.91** |
| Data.2 | SCRG | 4.62 | 15.33 | 14.04 | 20.79 | **23.79** | 12.29 | 5.93 | 19.31 | **26.83** |
|  | BSF | 1.99 | 5.03 | 7.06 | 8.69 | **8.71** | 7.86 | 7.08 | 6.92 | **9.95** |
| Data.3 | SCRG | 3.28 | 7.37 | 19.31 | 44.30 | **47.27** | 9.57 | 7.35 | 18.48 | **46.77** |
|  | BSF | 1.61 | 5.36 | 6.91 | 10.98 | **11.31** | 8.97 | 4.36 | 5.24 | **12.09** |
| Seq.1 | SCRG | 3.18 | 9.52 | 23.61 | **31.42** | – | 22.90 | 2.71 | 18.72 | **32.18** |
|  | BSF | 1.44 | 3.98 | 11.41 | **13.54** | – | 10.86 | 8.34 | 7.71 | **13.79** |
| Seq.2 | SCRG | 7.80 | 5.34 | 36.76 | **94.64** | – | 18.17 | 2.51 | 54.61 | **127.47** |
|  | BSF | 1.35 | 5.77 | 8.97 | **13.57** | – | 7.32 | 7.83 | 7.87 | **16.31** |
| Seq.3 | SCRG | 1.25 | 3.76 | 4.96 | 6.25 | **7.47** | 3.69 | 0.01 | 5.29 | **7.54** |
|  | BSF | 0.99 | 3.58 | 4.92 | **6.27** | 6.12 | 5.15 | 5.86 | 4.59 | **7.15** |
| Seq.4 | SCRG | 3.27 | 12.34 | 6.76 | **18.76** | – | 9.34 | 2.96 | 18.05 | **21.19** |
|  | BSF | 1.24 | 4.09 | 3.67 | **6.62** | – | 4.78 | 4.40 | 4.49 | **7.62** |
| Seq.5 | SCRG | 31.30 | 78.67 | 66.17 | 118.21 | 127.98 | 95.52 | 22.25 | **137.52** | **133.72** |
|  | BSF | 3.02 | 8.95 | 9.25 | 15.41 | **15.46** | 14.14 | 13.00 | 14.18 | **16.66** |

TABLE V
RUNTIME(S) OF NINE METHODS

| Test data | Top-hat | TLLCM | IPI | NRAM | RIPT | PSTNN | NTFRA | ANLPT | Proposed |
|---|---|---|---|---|---|---|---|---|---|
| Data.1 | 0.0067 | 2.3374 | 78.0669 | 11.8132 | 8.5549 | 0.4930 | 2.1744 | 2.6998 | 3.5457 |
| Data.2 | 0.0057 | 1.0340 | 5.2993 | 2.1900 | 1.8082 | 0.2097 | 1.2587 | 1.4200 | 1.4438 |
| Data.3 | 0.0113 | 2.3675 | 78.8006 | 12.3098 | 6.2475 | 0.3226 | 2.3139 | 2.8203 | 3.3652 |
| Seq.1 | 0.0040 | 1.0695 | 5.0153 | 1.4970 | 0.6250 | 0.0964 | 1.2261 | 1.3795 | 1.4652 |
| Seq.2 | 0.0049 | 1.0799 | 4.5079 | 1.4796 | 0.7352 | 0.1410 | 1.2193 | 1.4998 | 1.4472 |
| Seq.3 | 0.0039 | 1.0929 | 3.2239 | 2.0378 | 0.7937 | 0.1022 | 1.2271 | 1.3387 | 2.0215 |
| Seq.4 | 0.0038 | 1.0344 | 4.9506 | 2.0509 | 1.3601 | 0.1740 | 1.2447 | 1.3882 | 1.5547 |
| Seq.5 | 0.0188 | 0.9347 | 3.3013 | 1.1626 | 0.8898 | 0.1568 | 1.0099 | 1.0884 | 1.1199 |

*5) Algorithm Runtime:* TABLE V shows the average time spent in detection per frame of nine approaches. The experimental results show that Top-hat is much faster than HVS-based and low-rank and sparse decomposition-based models. It is worth noting that the tensor-based algorithms are faster than the matrix-based algorithms, which is more obvious when the image size is large. Among the low-rank and sparse decomposition-based algorithms, PSTNN has the best efficiency because of the early termination condition inside the algorithm, which may affect the detection performance when applied to complex scenes. Considering the excellent detection performance achieved by our method, the extra increase in runtime compared to the fastest PSTNN is acceptable.

## V. CONCLUSION

In this paper, in order to improve the performance of small target detection in complex scenes, a novel DWMGIPT model is proposed. By collecting nonoverlapping patches and tensor augmentation based on the TT decomposition, the MGIPT model is constructed. MGIPT obtains multi-granularity information from different modes, which helps us to approximate tensor rank more accurately. At the same time, the auto-weighted mechanism is introduced to measure the importance of TT rank of different modes and maintain more important information. Furthermore, steering kernel is used to extract prior information and suppress background interference. Finally, we provide an efficient iterative algorithm for solving the optimization function by applying ADMM. Compared with eight state-of-the-art approaches, extensive experimental results show that DWMGIPT can enhance the small target information effectively and suppress the clutter interference significantly in various complex scenes. Notably, the superiority and robustness of our algorithm are validated on lots of challenging scenes.

Since the proposed algorithm performs a large number of SVD calculations in the optimization process, the calculation cost is high. Therefore, in the next work, we will explore more efficient solution algorithm to improve real-time performance.

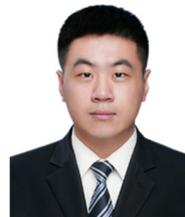
**Guiyu Zhang** received the B.E. degree in Mechanical Design manufacture and Automation from the Shandong University, China, in 2021. He is currently pursuing the M.S. degree in Electronic Information Engineering from University of Chinese Academy of Sciences (UCAS), China. His research interests focus on infrared target detection, image fusion and image super-resolution.

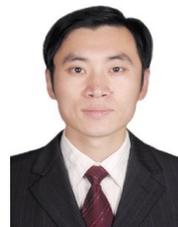
**Qunbo Lv** received the B.S. degree from Xidian University in 2001 and Ph.D. degree from Xi'an Institute of Optics and Precision Mechanics, Chinese Academy of Sciences (CAS) in 2007.

He works at the Key Laboratory of Computational Optical Imaging Technology, Aerospace Information Research Institute, CAS. He is a professor of University of the Chinese Academy of Sciences (UCAS). His primary research fields include optical imaging technology, spectral imaging technology and image processing.

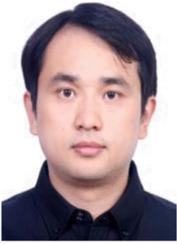

**Zui Tao** was born in China in 1984. He received the B.S. degree from Henan University, Kaifeng, China, in 2005, the M.S. degree from Wuhan University, Wuhan, China, in 2008, and the Ph.D. degree from the Institute of Remote Sensing and Digital Earth (RADI), Chinese Academy of Sciences (CAS), Beijing, China, in 2012, all in cartography and geographic information system.

He is currently an Associate Professor with the Aerospace Information Research Institute, CAS. His research interests include the validation of remote sensing product, ecological, and environmental remote sensing.

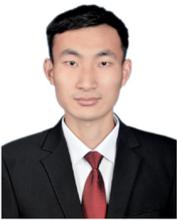

**Baoyu Zhu** received the B.E. degree in Mechanical Design manufacture and Automation from the Shandong University, China, in 2020, and the M.S. degree in Electronic Information Engineering from University of Chinese Academy of Sciences (UCAS), China, in 2023. He is currently pursuing the Ph.D. degree at UCAS, China, with research interests in object detection and image processing.

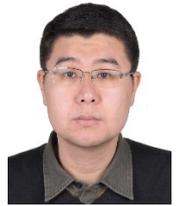

**Zheng Tan** received the B.E. and M.E. degrees from Northwestern Polytechnical University in 2005 and 2009 respectively, and the Ph.D. degree from University of the Chinese Academy of Sciences in 2018. Since 2018, he has been an associate professor of engineering with the Aerospace Information Research Institute, Chinese Academy of Sciences. His research interests include remote sensing data processing and computer vision.

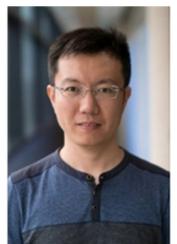

**Yuan Ma** received the B.E. degree from the University of Electronic Science and Technology of China and the Ph.D. degree from the University of Chinese Academy of Sciences in 2011 and 2017, respectively. He is an assistant professor with the Department of Mechanical Engineering, at Tsinghua University, and his research interests are in micro-nano manufacture technology.